\documentclass[sigconf]{acmart}

\usepackage{epstopdf}
\usepackage{booktabs} 
\usepackage{amsmath}
\usepackage{amssymb}
\usepackage[ruled,vlined]{algorithm2e}
\usepackage{multirow}

\usepackage{helvet,url}
\usepackage{courier}
\usepackage{epsfig}
\usepackage{picinpar}
\usepackage{graphicx}
\usepackage{epstopdf}
\usepackage{amsmath}
\usepackage{amsthm}
\usepackage{amssymb,bm}
\usepackage[ruled,vlined]{algorithm2e}
\usepackage[titletoc]{appendix}
\usepackage{float,caption}
\usepackage{epsfig}
\usepackage{picinpar}
\usepackage{amsmath}
\usepackage{amsthm}
\usepackage{amssymb,bm}
\usepackage[ruled,vlined]{algorithm2e}
\usepackage[titletoc]{appendix}
\usepackage{float,caption,multirow}
\usepackage{subcaption,xcolor}

\newcommand{\bs}{\mathbf}

\newcommand{\mb}{\mathbb}

\copyrightyear{} 
\acmYear{} 
\setcopyright{none}

\acmConference{Manuscript}{2018}{ArXiv}\acmPrice{}\acmDOI{}
\acmISBN{}

\begin{document}

\title{Fast Single Image Rain Removal via a \\Deep Decomposition-Composition Network}

\author{Siyuan Li$^{1}$, Wenqi Ren$^{2}$, Jiawan Zhang$^{1}$, Jinke Yu$^{3}$ and Xiaojie Guo$^{1}$ }
\affiliation{
	$^1$ Tianjin University\quad $^2$ IIE, Chinese Academy of Sciences \quad $^3$ Dalian University of Technology}
%
%

%
%


\begin{abstract}
Rain effect in images typically is annoying for many multimedia and computer vision tasks. For removing rain effect from a single image, deep leaning techniques have been attracting considerable attentions. This paper designs a novel multi-task leaning architecture in an end-to-end manner to reduce the mapping range from input to output and boost the performance. Concretely, a decomposition net is built to split rain images into clean background and rain layers. Different from previous architectures, our model consists of, besides a component representing the desired clean image, an extra component for the rain layer. During the training phase, we further employ a composition structure to reproduce the input by the separated clean image and rain information for improving the quality of decomposition. Experimental results on both synthetic and real images are conducted to reveal the high-quality recovery by our design, and show its superiority over other state-of-the-art methods. Furthermore, our design is also applicable to other layer decomposition tasks like dust removal. More importantly, our method only requires about 50ms, significantly faster than the competitors, to process a testing image in VGA resolution on a GTX 1080 GPU, making it attractive for practical use.
\end{abstract}

%
%
%



\maketitle

\section{Introduction}
\begin{figure}[t]
	\begin{center}
		\begin{subfigure}{0.48\linewidth}
			\includegraphics[width=1\linewidth]{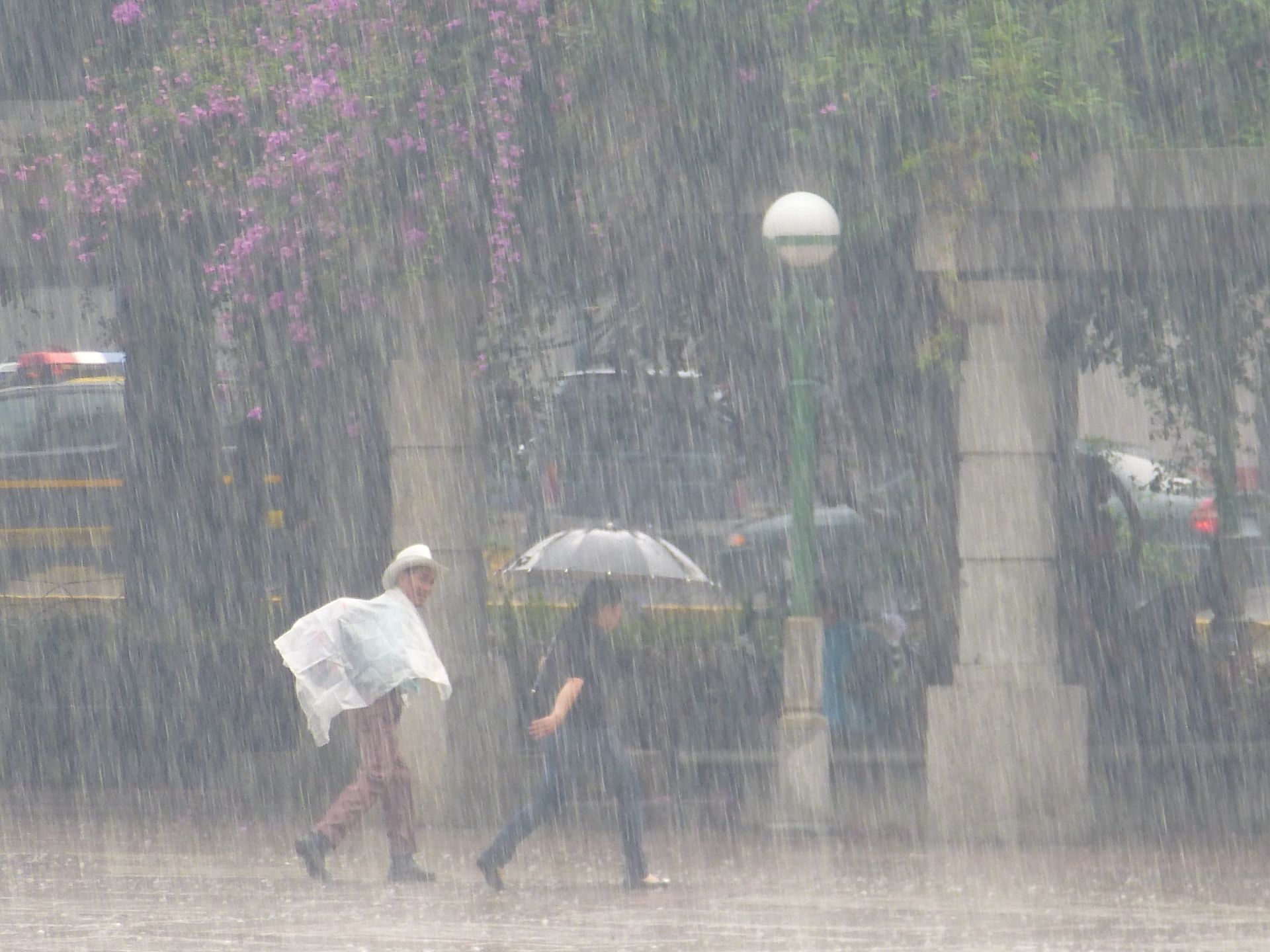}
		\end{subfigure}
		\begin{subfigure}{0.48\linewidth}
			\includegraphics[width=1\linewidth]{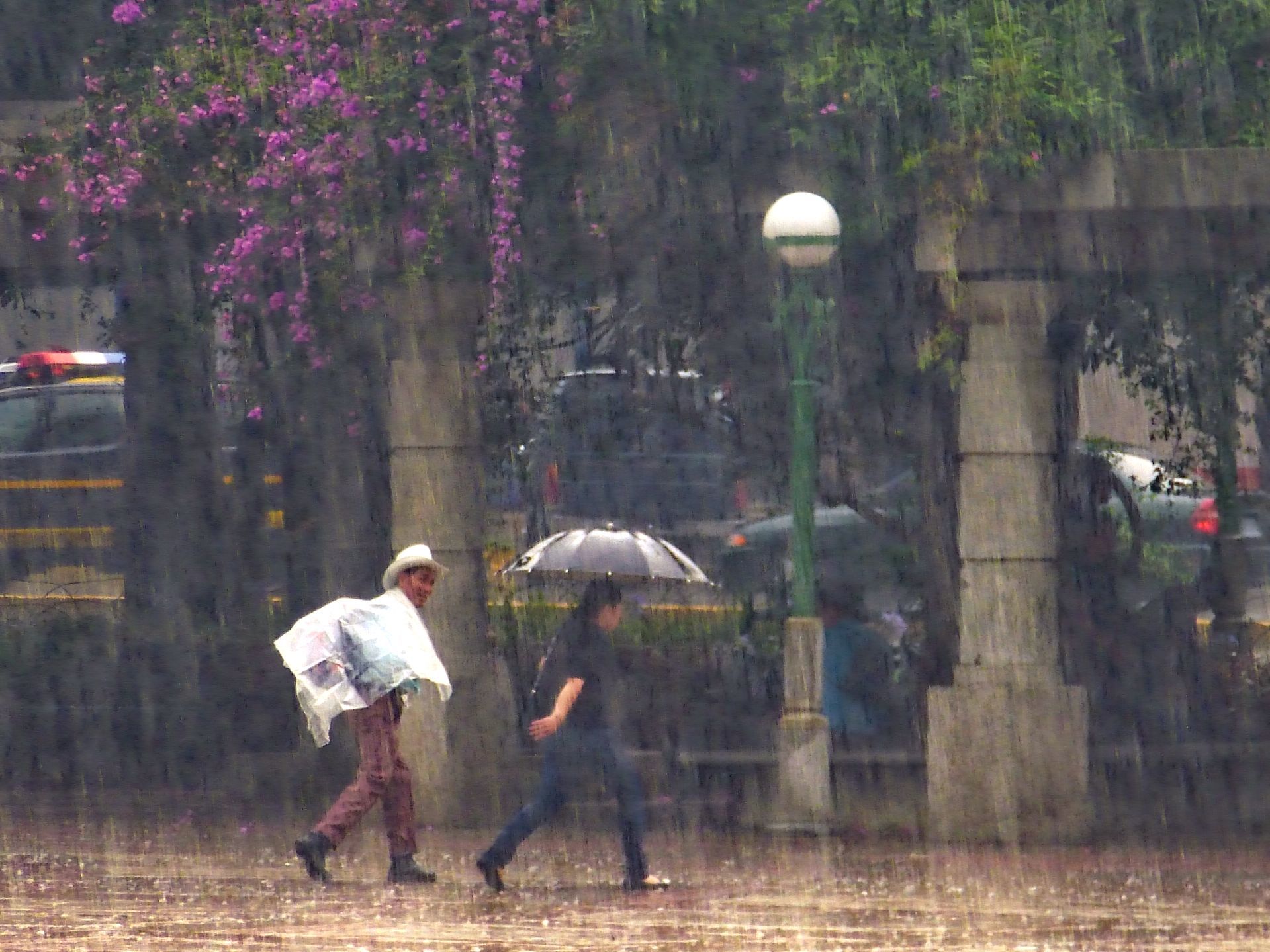}
		\end{subfigure}\\ \vspace{3pt}
		
		\begin{subfigure}{0.48\linewidth}
			\includegraphics[width=1\linewidth]{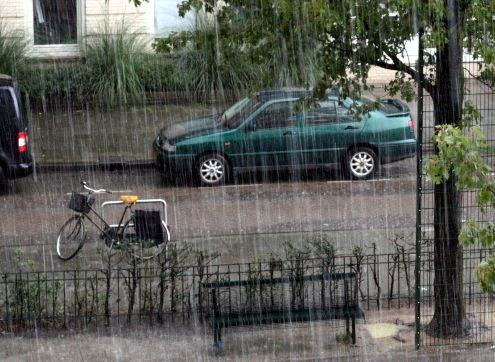}
		\end{subfigure}
		\begin{subfigure}{0.48\linewidth}
			\includegraphics[width=1\linewidth]{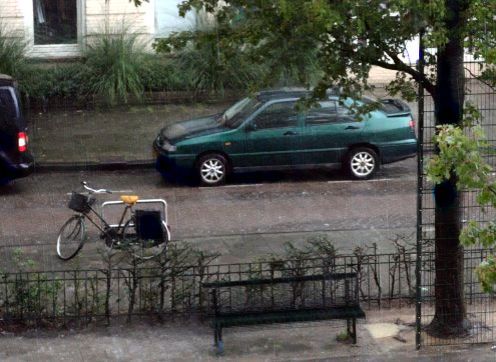}
		\end{subfigure}
	\end{center}
	\caption{Left column: Two natural images interfered by rains. Right column: Corresponding rain removed results by our method. The background details are very well preserved while the rain effects are greatly reduced/eliminated.}
	\label{fig:open}
\end{figure}

Most, if not all, of classic and contemporary vision-oriented algorithms, such as object detection \cite{objDet} and tracking \cite{Tracking}, can work reasonably well when facing images of high visibility, but dramatically degenerate or even fail if they are fed with low-quality inputs. In real-world scenarios, especially for outdoor scenes, rain effect has always been such an annoying and inevitable nuisance, which would significantly alter or degrade the content and color of images \cite{Effect}. These situations frequently occur if one records an event happening at a square using a smart phone, a surveillance camera monitors a street, or an autonomous vehicle drives on a road, in rainy days. The rain in atmosphere generally has two existence styles, say steady rain and dynamic rain. The steady rain is caused by distant microscopic rain drops globally accumulated throughout the scene, while the dynamic one comes from large particles (rain streaks) that look like random and local corruptions. The left column of Fig. \ref{fig:open} gives two such examples.  For eliminating or reducing negative effects brought by rain, the development of effective approaches is demanded.   

Formally, the rainy image can be seen as a superimposition of two layers $\bs{O}= \Psi(\bs{B}, \bs{R}$), where $\bs{O}\in{\mb{R}^{m\times n}}$ designates the observed data and, $\bs{R}\in\mb{R}^{m\times n}$ and $\bs{B}\in\mb{R}^{m\times n}$ represent the rain layer and the desired clean background, respectively. In addition, $\Psi(\cdot,\cdot)$ is a blending function. To decompose the two layers from a single image is mathematically ill-posed since the number of unknowns to recover is twice as many as the given measurements. 

\subsection{Previous Arts and Challenges}
Over the past decades,  a lot of attentions to resolving the rain removal problem have been drawn from the community.
From the perspective of required input amount, existing rain removal methods can be divided into two classes, \textit{i.e.} multi-image based and single image based methods. Early attempts on deraining basically belong to the former category. A representative solution was proposed in \cite{GargC}, based on the recognition that the visibility of rain in images depends much on the exposure time and depth of field of the camera. One can achieve the goal by testing on several images and adjusting the operational parameters of the camera. {But this method is too professional to use for typical consumers.} The work in \cite{GargM2} employs two constraints to automatically find and exclude possible rain streaks, and then fills up the holes by averaging the values of their temporal neighbors, which releases the professional requirement. Several follow-ups along this technique line including \cite{ZhangM} and \cite{YouM} try to improve the accuracy of rain streak detection or/and the quality of background inpainting. A more elaborated review on the multi-image based rain streak removal approaches can be found in \cite{ReviewM}. \emph{Generally, this kind of methods can provide reasonable results when the given information is of sufficient redundancy, but this condition is often violated in practice.} 

For the sake of flexibility and applicability, single image based approaches are more desirable but challenging. Kang \textit{et al.} \cite{KangS} proposed a two-step method. The first step is separating the input rain image into a low-frequency component containing its structure and a high-frequency one with both rain streaks and background textures. Then the image textures are distinguished from the rain streaks in the detail layer according to constructed dictionaries, and added back to the structure layer. However, the separation in the detail layer is challenging, always tending to either over-smooth the background or leave noticeable rain steaks. 
Its follow-ups include \cite{HuangS,SunS}. Chen and Hsu \cite{ChenS} proposed a unified objective function for rain removal by exploring the repetitive property of the rain streak appearance and using a low rank model to regularize the rain streak layer. This is problematic as other repetitive structures like building windows also fit the low-rank assumption. Kim \textit{et al.} \cite{KimS} tried to detect rain streaks by a kernel regression method, and remove the suspects via a nonlocal mean filtering. 
It frequently suffers from inaccurate detection of rain streaks. Luo \textit{et al.} \cite{LuoS} created a new blending model and attempted to reconstruct the background and rain layers of image patches over a self-trained dictionary by discriminative sparse coding. Although the method has an elegant formulation, the blending model used still needs physical validation and the effectiveness in removing the rain is somehow weak as one can always see remaining thin structure at the rain streak locations in the output. Li \textit{et al.} \cite{LiS} used patch-based priors for both the two layers, namely a \textit{Gaussian mixture model} (GMM) learned from external clean natural images for the background and another GMM trained on rain regions selected from the input image itself for the rain layer. \emph{These prior-based methods even with the help of trained dictionaries/GMMs, on the one hand, are still unable to catch sufficiently distinct features for the background and rain layers. On the other hand, their computational cost is way too huge for practical use.}

With the emergence of deep learning, a number of low-level vision tasks have benefited from deep models supported by large-scale training data, such as \cite{NIPSDen,TIPDen} for denoising, \cite{PAMISR} for super-resolution, \cite{ICCVCmp} for compression artifact removal and \cite{DehNet} for dehazing, as the deep architectures can better capture explicit and implicit features. As for deraining, Fu \textit{et al.} proposed a \textit{deep detail network} (DDN) \cite{DDN}, inspired by \cite{KangS}. It first decomposes a rain image into a detail layer and a structure layer. Then the network focuses on the high-frequency layer to learn the residual map of rain streaks. The restored result is formed by adding the extracted details back to the structure layer. Yang \textit{et al.} \cite{JORDER} proposed a \textit{convolutional neural network} (CNN) based method to \textit{jointly detect and remove rain streaks} from a single image (JORDER). They used a multi-stream network to capture the rain streak component with different scales and shapes. The rain information is then fed into the network to further learn rain streak intensity. By recurrently doing so, the rain effect can be detected and removed from input images.  The work in \cite{ID-CGAN} proposes a single image de-raining method called \textit{image deraining conditional general adversarial network} (ID-CGAN), which considers quantitative, visual and also discriminative performance into the objective function. \emph{Though the deep learning based  strategies have made a great progress in rain removal compared with the traditional methods, two challenges still remain:} 
\begin{itemize}
	\item \emph{How to enhance the effectiveness of deep architectures for better utilizing training data and achieving more accurate restored results;}
	\item \emph{How to improve the efficiency of processing testing images for fulfilling the high-speed requirement in real-world (real-time) tasks.}
\end{itemize}

\subsection{Our Contributions}
\begin{figure*}[t]
	\begin{center}
		\includegraphics[width=.95\linewidth]{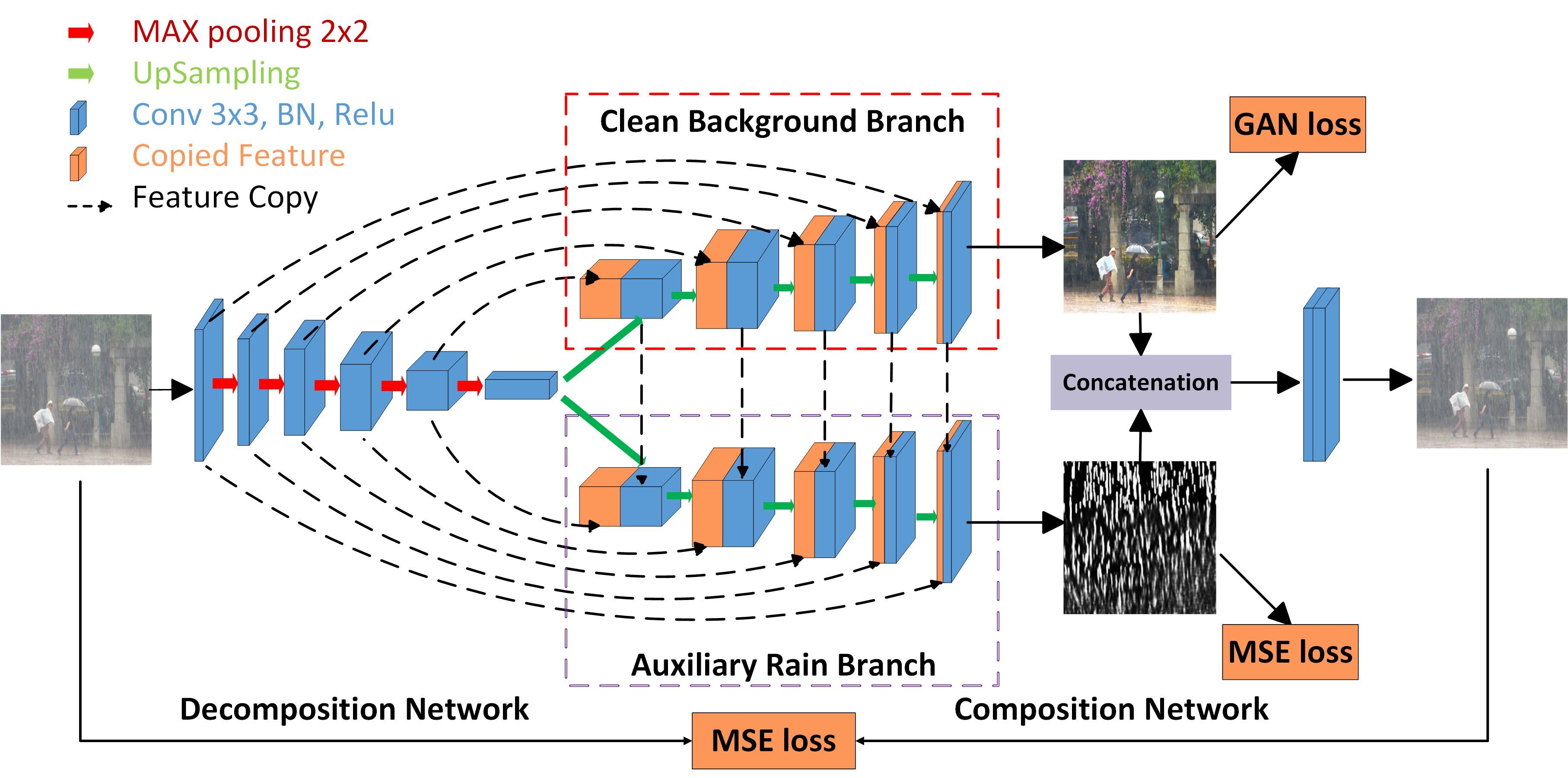}
	\end{center}
	\caption{Architecture of our proposed Deep Decomposition-Composition Network}
	\label{fig:framework}
\end{figure*}

In order to address the aforementioned challenges, we propose a novel deep decomposition-composition network (DDC-Net) to effectively and efficiently remove the rain effect from a single image under various conditions. Concretely, the contributions can be summarized as follows. \emph{The designed network is composed by a decomposition net and a composition net. The decomposition net is built for splitting rainy images into clean background and rain layers. The volume of model is retained small with promising performance. Hence, the effectiveness of the architecture is boosted. The composition net is for reproducing input rain images by the separated two layers from the decomposition net, aiming to further improve the quality of decomposition. Different from previous deep models, ours explicitly takes care of the recovery accuracy of the rain layer. According to the screen blending mode, instead of the simple additive blending, we synthesize a training dataset containing $10,400$ triplets [rain image, clean background, rain information]. During the testing phase, only the decomposition net is needed. Experimental results on both synthetic and real images are conducted to reveal the high-quality recovery by our design, and show its superiority over other state-of-the-art methods. Our method is significantly faster than the competitors, making it attractive for practical use. All the trained models and the synthesized dataset are available at}\\ \url{https://sites.google.com/view/xjguo}. 

\section{Deep Decomposition-Composition Network}

The designed network architecture is illustrated in Fig. \ref{fig:framework}. It consists of two modules, \textit{i.e.}, the decomposition network and the composition network. 

\subsection{Decomposition Net}
As can be seen from Fig. \ref{fig:framework}, the decomposition network, aiming to separate the rain image into the clean background and rain layers, has two main branches: one focuses on restoring the background and the other for the rain information. 

Inspired by the effectiveness of encoder-decoder networks in image denoising~\cite{mao2016image}, inpainting~\cite{pathak2016context} and matting~\cite{xu2017deep}, we construct our decomposition branch based on the residual encoder and decoder architecture with specific designs for clean background and rain layer prediction as follows: 1) the first two convolutional layers in the encoder are changed to dilated convolution~\cite{yang2017deep} to enlarge the receptive field. The stride of our dilated convolutional layer is $1\times 1$ with padding. We use max-pooling to down-sample feature maps; 
2) we use two decoder networks (clean background branch and auxiliary rain branch) to recover the clean background and rain layer respectively;
and 3) features from the deconvolution module of the clean background branch are concatenated to the auxiliary rain branch for better obtaining rain information during the up-sampling stage (the downward arrows). The principle behind is that the background features are expected to help exclude textures belonging to the background from the rain part.\footnote{ It is worth to note that we have tried to feed rain features into the background branch, but this operation did not show noticeable improvement on performance. The reason is that since the rain textures are typically much simpler and more regular than the background textures, the improvement is not obvious or negligible.  }

The residual encoder and decoder network has five convolution modules, and each consists of several convolutional layers, ReLu and skip links. Specifically, the feature maps from the 1$st$, 2$nd$, 3$rd$, 4$th$ and 5$th$ convolution modules are with sizes of 1/2, 1/4, 1/8, 1/16 and 1/32 of the input image size, respectively. The corresponding decoder introduces up-sampling operation to build feature maps up from low to high resolution. \\

\noindent\textbf{Pre-train on synthetic images:}
Since the decomposing problem is challenging without paired supervision,
a model can learn arbitrary mapping to the target domain and
cannot guarantee to map an individual input to its desired
clean background and rain layers. 
Therefore, we first cast the image deraining problem as a paired image-to-image mapping problem, in which we use the clean images and the corresponding synthesized rainy image as the paired information. To measure the content difference between the recovered and the ground truth image, we use the Euclidean distance between the obtained results and target images. Therefore, the losses of clean background and rain layer in the decomposition in the pre-training stage can be written as:
\begin{equation}
\mathcal{L}_\bs{B} = \frac{1}{N}\sum_{i=1}^{N}\big\|f_b(\bs{O}^i)-\bs{B}^i\big\|^2_F,
\label{eq-loss-mse1}
\end{equation}
and
\begin{equation}
\mathcal{L}_\bs{R} = \frac{1}{N}\sum_{i=1}^{N}\big\|f_r(\bs{O}^i)-\bs{R}^i\big\|^2_F,
\label{eq-loss-mse2}
\end{equation}
where $N$ denotes the number of training images in each process, $\|\cdot\|_F$ designates the Frobenius norm and, $\bs{O}^i$, $\bs{B}^i$ and $\bs{R}^i$ stand for the i$th$ input, background and rain images, respectively. In addition, $f_b$ and $f_r$ are the clean background branch and auxiliary rain branch, respectively. \\

\noindent\textbf{Fine-tune on real images:} With the learned model by using synthesized images, our decomposition network could guarantee to map an individual rainy input to its desired clean background and rain layer. But, as the synthetic rain layer cannot distinguish the effect of attenuation and splash in real scene radiance, we use some collected real rain-free and rainy images to fine-tune our model.

We propose to solve the unpaired image-to-image problem by introducing the \textit{generative adversarial network} GAN loss \cite{goodfellow2014generative,yang2018towards} to better model the formation of rainy images. The concept of GAN was first proposed by Goodfellow \emph{et al.} \cite{goodfellow2014generative}, which has been attracting substantial attention from the community. The GAN is based on the minimax two-player game, which can provide a simple yet powerful way to estimate target distribution and generate new samples. The GAN framework consists of two adversarial models: a generative model $G$ and a discriminative model $D$. The generative model $G$ can capture the data distribution, and the discriminative model $D$ can estimate the probability that a sample comes from the training data rather than $G$. The generator takes the noise as input and tries to generate different samples to fool the discriminator, and the discriminator aims to determine whether a sample is from the model distribution or the data distribution, finally the generator generates samples that are not distinguishable by the discriminator. 

In this stage, we do not need any supervision for rain layers, but for the purpose of promoting the performance to be realistic, on the account of learning a generative adversarial network via distinguishing the real image and the fake image to simulate the restored images, we hope the adversarial model can assist to train the decomposition network that generates enhanced images which can cheat the discriminator $D$ to distinguish from real images. The adversarial loss is defined as follows.
\begin{eqnarray}
\begin{split}
\mathcal{L}_{\text{ADV}} = & \underset{\bs{I}\backsim p(\mathcal{I})}{\mathbb{E}}\big[\log D(\bs{I})\big] \\ &
+\underset{\bs{O}\backsim p(\mathcal{O})}{\mathbb{E}}\left[\log\big(1-D(G(\bs{O}))\big)\right],
\end{split}
\label{equ-loss-adv}
\end{eqnarray}
The discriminator $D$ consists of  five convolutional layers each followed by a ReLU nonlinearity. The detailed configuration of $D$ is given in Table~\ref{tab-adv}. A sigmoidal activation function is applied to the outputs of the last convolutional layer for producing a probability of the input image being detected as ``real'' or ``fake''.


%
%
\begin{table}[t]
	\caption{Model parameters of the discriminator. Every convolution layers are activated with a LeakyReLU layer.}

	\begin{center}\small{
			\begin{tabular}{c||c||c||c}
				\hline
				Layer & Kernel dimension  & Stride   &  Output size   \\
				\hline\hline
				Input & - & - &  $224\times 224$\\
				Conv1 & $64\times 4\times 4$   & 2 &  $112\times 112$\\
				Conv2 & $128\times 4\times 4$   & 2 &  $56\times 56$\\
				Conv3 & $256\times 4\times 4$   & 2 &  $28\times 28$\\
				Conv4 & $512\times 4\times 4$   & 1 &  $27\times 27$\\
				Conv5 & $1\times 4\times 4$   & 1 &  $26\times 26$\\
				Sigmoid & - & - & -\\
				\hline            
		\end{tabular}}
		\label{tab-adv}
	\end{center}
	
\end{table}

\subsection{Composition Net}

Our composition net aims to learn the original rainy image from the outputs of decomposition model, then use the constructed rainy image as the self-supervised information to guide the back-propagation.
With the decomposition model, we can disentangle a rainy image
into two corresponding components.
The first one is the recovered rain-free image $\bs{B}$ from the clean background branch. The second one is the rain layer, denoted as $\bs{R}$, learned by the auxiliary rain branch. Therefore, we can directly compose the corresponding rain image $\bs{O}$ in a simple way:
\begin{equation}
\bs{O} = \bs{B} + \bs{R}.
\label{eq1}
\end{equation}
Although Eq. \eqref{eq1} is a standard model in the de-raining literature \cite{xu2017deep}, it is limited in real cases since there are other factors in rainy images, such as haze and splashes.

To solve this problem, we first concatenate the clean background image and the rain layer from the decomposition network, and then adopt an additional CNN block to model the real rainy image formulation. The proposed composition network could achieve a more general formation process and accounts for some unknown phenomenon in real images.  
Then we define a quadratic training cost function to
measure the difference between the reconstruct rainy output and the
original rainy image as
\begin{equation}
\mathcal{L}_{\bs{O}} = \frac{1}{N}\sum_{i=1}^{N}\big\|f(\bs{O})-\bs{O}\big\|^2_F.
\label{eq-loss-mse3}
\end{equation}
where $f$ is the whole network depicted in Fig. \ref{fig:framework}. We here notice that, in the testing stage, only the clean background branch is required to generate desired results .\\

\subsection{Training Dataset}
We notice that, instead of the additive mode ($\bs{O}=\bs{B}+\bs{R}$), we advocate the screen blend one to synthesize data for better approximating real-word cases. The screen mode is in the form:
\begin{equation}
\begin{aligned}
\bs{O} = &\bs{1}-(\bs{1}-\bs{B})\circ(\bs{1}-\bs{R})\\
=&\bs{B}+\bs{R}-\bs{B}\circ\bs{R},
\end{aligned}
\end{equation} 
where $\circ$ means the Hadamard product.
In this mode, the values of the pixels in the two layers are inverted, multiplied, and then inverted again. The clean background images $\bs{B}$ are from BSD300 dataset\footnote{\url{https://www2.eecs.berkeley.edu/Research/Projects/CS/vision/bsds/}}. Moreover, the rain part $\bs{R}$ is generated following the steps\footnote{\url{http://www.photoshopessentials.com/photo-effects/rain/}} with varying intensities, orientations and overlaps. Finally, we fuse the two parts in the way of screen blending. In this manner, the synthesized rainy images are more physically meaningful. As a consequence, we obtain a dataset containing $10,400$ triplets [rainy image, clean background, rain layer]. As for the fine-tune stage, we collect $15$ real images and randomly crop $240$ samples with the size of $224 \times 224$ from them.  

\subsection{Implementation Details}


During training we use a batch size of $8$, and patch size of $224 \times 224$. 
We use the stochastic gradient descent (SGD) technique for optimization with the weight decay of $10^{-6}$ and momentum of $0.9$. The maximum number of iterations is 100K and we adopt the learning rate of $10^{-3}$ for the first 70K iterations, then continue to train for 30K iterations with the learning rate of $10^{-4}$. The entire network is trained on a Nvidia GTX 1080 GPU using the Keras framework.
During the testing phase, only the background decomposition branch is used, which is efficient.

\section{Experimental Verification}
This section evaluates our DDC-Net on the task of rain removal, in comparison with the state-of-the-arts including GMM \cite{LiS}, JORDER \cite{JORDER}, DDN \cite{DDN} and ID-CGAN \cite{ID-CGAN}. The codes are either downloaded from the authors' websites or provided by the authors. For quantitatively measuring the performance, we employ PSNR, SSIM and elapsed time as the metrics. All the comparisons shown in this paper are conducted under the same hardware configuration.

\begin{figure*}[t]
	\begin{center}
		\begin{subfigure}{0.16\linewidth}
			\includegraphics[height=2cm]{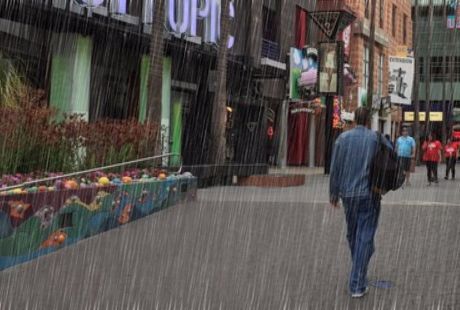}
		\end{subfigure}
		\begin{subfigure}{0.15\linewidth}
			\includegraphics[height=2cm]{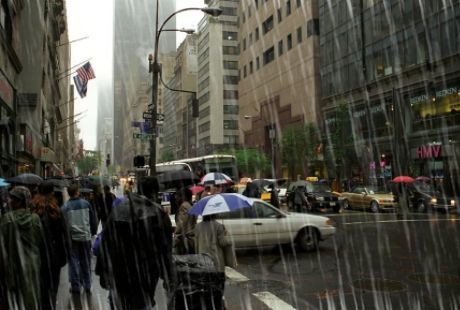}
		\end{subfigure}
		\begin{subfigure}{0.10\linewidth}
			\includegraphics[height=2cm]{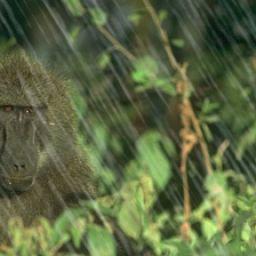}
		\end{subfigure}
		\begin{subfigure}{0.13\linewidth}
			\includegraphics[height=2cm]{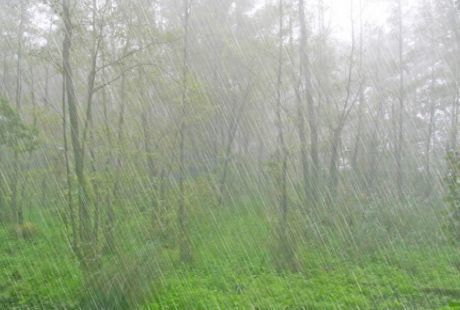}
		\end{subfigure}
		\begin{subfigure}{0.13\linewidth}
			\includegraphics[height=2cm]{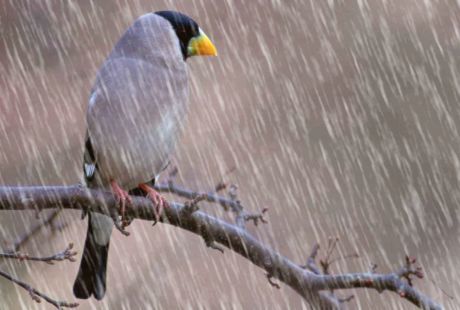}
		\end{subfigure}
		\begin{subfigure}{0.13\linewidth}
			\includegraphics[height=2cm]{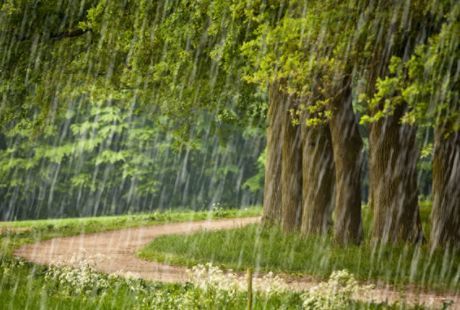}
		\end{subfigure}
		\begin{subfigure}{0.13\linewidth}
			\includegraphics[height=2cm]{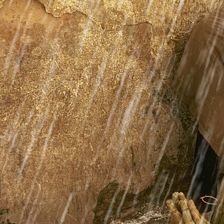}
		\end{subfigure}
	\end{center}
	
	\rule{\textwidth}{0.5pt}\\
	
	\begin{center}
		\begin{subfigure}{0.135\linewidth}
			\includegraphics[width=1\linewidth]{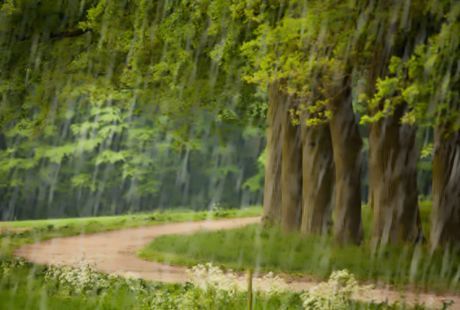}
		\end{subfigure}\hspace{-1pt}
		\begin{subfigure}{0.135\linewidth}
			\includegraphics[width=1\linewidth]{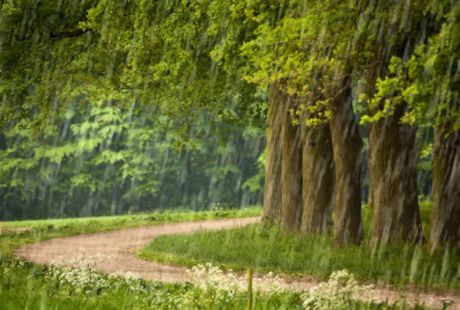}
		\end{subfigure}\hspace{-1pt}
		\begin{subfigure}{0.135\linewidth}
			\includegraphics[width=1\linewidth]{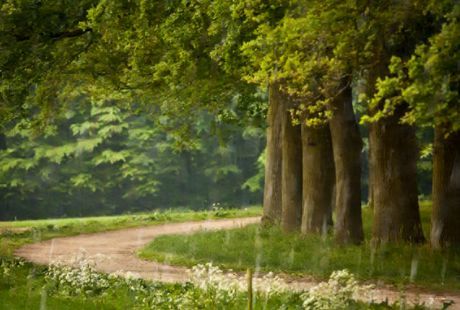}
		\end{subfigure}\hspace{-1pt}
		\begin{subfigure}{0.135\linewidth}
			\includegraphics[width=1\linewidth]{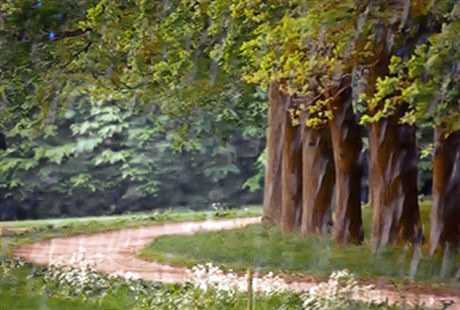}
		\end{subfigure}\hspace{-1pt}
		\begin{subfigure}{0.135\linewidth}
			\includegraphics[width=1\linewidth]{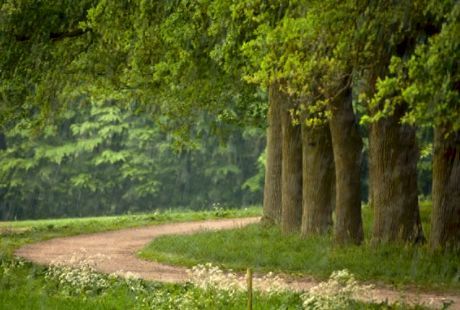}
		\end{subfigure}\hspace{-1pt}
		\begin{subfigure}{0.135\linewidth}
			\includegraphics[width=1\linewidth]{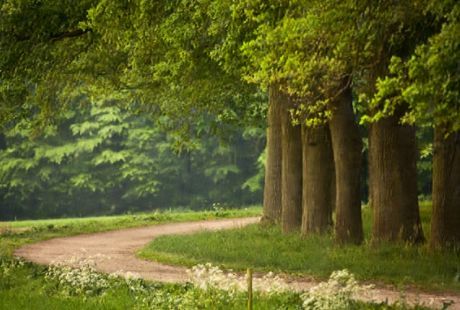}
		\end{subfigure}\hspace{-1pt}
		\begin{subfigure}{0.135\linewidth}
			\includegraphics[width=1\linewidth]{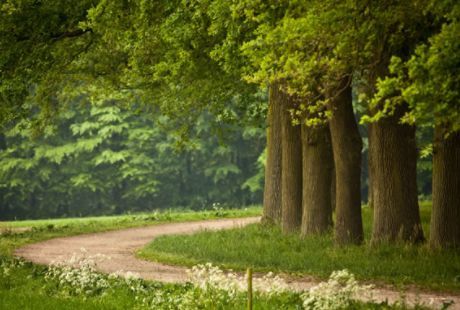}
		\end{subfigure}\\ \vspace{3pt}
		
		\begin{subfigure}{0.135\linewidth}
			\includegraphics[width=1\linewidth]{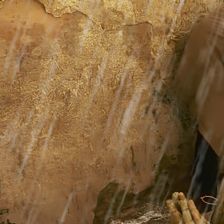}
			\subcaption{GMM}
		\end{subfigure}\hspace{-1pt}
		\begin{subfigure}{0.135\linewidth}
			\includegraphics[width=1\linewidth]{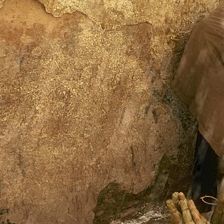}
			\subcaption{JORDER}
		\end{subfigure}\hspace{-1pt}
		\begin{subfigure}{0.135\linewidth}
			\includegraphics[width=1\linewidth]{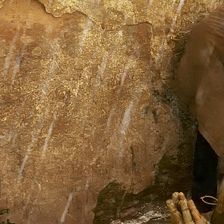}
			\subcaption{DDN}
		\end{subfigure}\hspace{-1pt}
		\begin{subfigure}{0.135\linewidth}
			\includegraphics[width=1\linewidth]{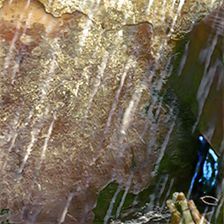}
			\subcaption{ID-CGAN}
		\end{subfigure}\hspace{-1pt}
		\begin{subfigure}{0.135\linewidth}
			\includegraphics[width=1\linewidth]{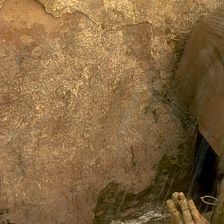}
			\subcaption{Ours w/o CN}
		\end{subfigure}\hspace{-1pt}
		\begin{subfigure}{0.135\linewidth}
			\includegraphics[width=1\linewidth]{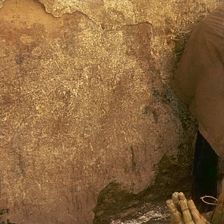}
			\subcaption{Our DDC}
		\end{subfigure}\hspace{-1pt}
		\begin{subfigure}{0.135\linewidth}
			\includegraphics[width=1\linewidth]{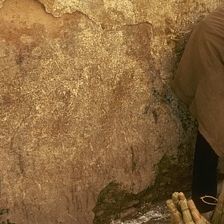}
			\subcaption{GrounTruth}
		\end{subfigure}\\
	\end{center}
	\caption{Top row shows the synthetic images used in the quantitative evaluation. The rest rows give two comparisons.}
	\label{fig:synt}
\end{figure*}

\begin{table}[t]
	\caption{Performance comparison in terms of PSNR and SSIM. The best results are highlighted in bold. }
	\begin{center}
		\resizebox{0.48\textwidth}{!}{
			\begin{tabular}{ c|| c c c c c c }
				\hline
				\textbf{Method} & \textbf{GMM}  & \textbf{JORDER} & \textbf{DDN} & \textbf{ID-CGAN} & \textbf{Our w/o CN} & \textbf{Our DDC}  \\
				\hline
				\hline 
				PSNR & 30.5659 & 29.6848 & \textbf{32.8732} & 19.9845 & 29.6728 & 32.5450\\
				SSIM  & 0.8846 & 0.8686  & \textbf{0.9184} & 0.7965 & 0.8500 &  0.9113 \\
				\hline
				\hline
				PSNR & 30.1166 & 31.5563 & 31.7156 & 22.7293 & 31.1119 &\textbf{34.1597} \\
				SSIM  & 0.8600 & 0.9198 & 0.9094 & 0.7507  & 0.8565 & \textbf{0.9223} \\
				\hline
				\hline 
				PSNR & 29.8729 & 28.8450 & 30.8381 & 23.6769 & 35.0035 & \textbf{35.6594}\\
				SSIM  & 0.8516 & 0.8682  & 0.9189 & 0.6662 & 0.9371 & \textbf{0.9535} \\
				\hline
				\hline
				PSNR & 33.8396 & 33.3301 & 37.0646 & 19.4511 & 31.5210 &\textbf{38.3306} \\
				SSIM  & 0.8425 & 0.9228 & 0.9326 & 0.7345  & 0.8633 & \textbf{0.9408} \\
				\hline
				\hline
				PSNR & 30.5819 & 32.2956 & 35.1259 & 20.1831 & 33.2226 &\textbf{36.0266} \\
				SSIM  & 0.9047 & 0.9290 & \textbf{0.9490} & 0.5902  & 0.8980 & {0.9464} \\
				\hline
				\hline
				PSNR & 25.7104 & 24.7909 & 30.3679 & 22.9528 & 31.0195 &\textbf{33.0215} \\
				SSIM  & 0.6892 & 0.7627 & 0.8573 & 0.6711  & 0.8520 & \textbf{0.8867} \\
				\hline
				\hline
				PSNR & 28.6585 & 31.4446 & 29.8325 & 21.3416 & 34.8412 &\textbf{38.5638} \\
				SSIM  & 0.7838 & 0.9153 & 0.8890 & 0.6402  & 0.9225 & \textbf{0.9653} \\
				\hline
			\end{tabular}
		}
	\end{center}
	
	\label{tab:synm}
\end{table}

\begin{table*}[t]
	\caption{Running time comparison (seconds). The best results are highlighted in bold.  }
	\begin{center}
		\begin{tabular}{ c|| c c c c c c }
			\hline
			\textbf{Image Size} & \textbf{GMM} & \textbf{JORDER}  & \textbf{DDN} & \textbf{ID-CGAN} & \textbf{Our DDC} \\
			\hline
			\hline 
			250$\times$250 & 234.9 (CPU)& 48.59/2.23 (CPU/GPU) & 1.81/0.27 (CPU/GPU)& 0.15 (GPU)& \textbf{0.98}/\textbf{0.03} (CPU/GPU)\\
			500$\times$500  & 772.4 (CPU)& 88.61/3.34  (CPU/GPU)& 13.27/0.74 (CPU/GPU)& 0.55 (GPU)& \textbf{4.04}/\textbf{0.12} (CPU/GPU)\\
			\hline
		\end{tabular}
	\end{center}

	\label{tab:time}
\end{table*}

\begin{figure*}[t]
	\begin{center}
		\begin{subfigure}{0.33\linewidth}
			\includegraphics[width=1\linewidth,height=4cm]{3.jpg}
		\end{subfigure}
		\begin{subfigure}{0.33\linewidth}
			\includegraphics[width=1\linewidth,height=4cm]{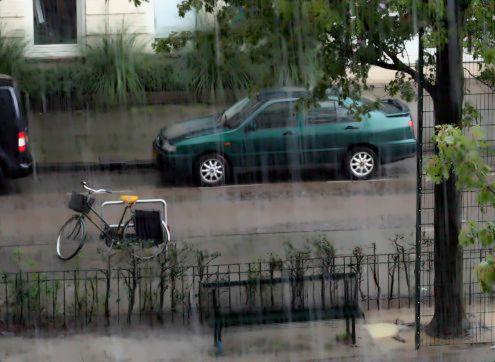}
		\end{subfigure}
		\begin{subfigure}{0.33\linewidth}
			\includegraphics[width=1\linewidth,height=4cm]{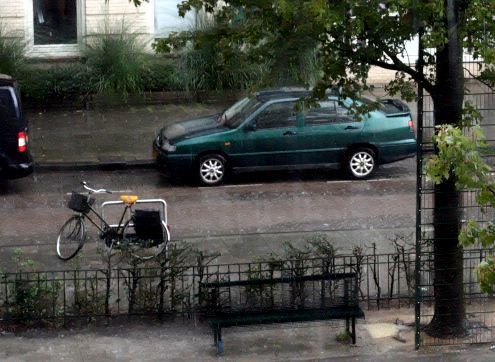}
		\end{subfigure}\\ \vspace{2pt}
		
		\begin{subfigure}{0.33\linewidth}
			\includegraphics[width=1\linewidth,height=4cm]{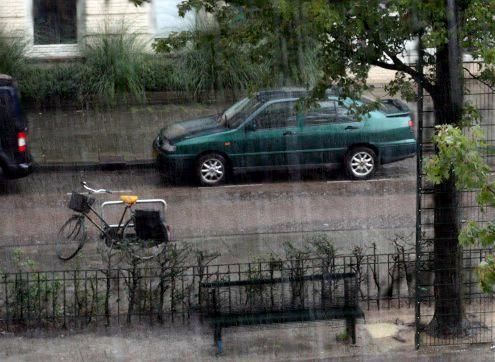}
		\end{subfigure}
		\begin{subfigure}{0.33\linewidth}
			\includegraphics[width=1\linewidth,height=4cm]{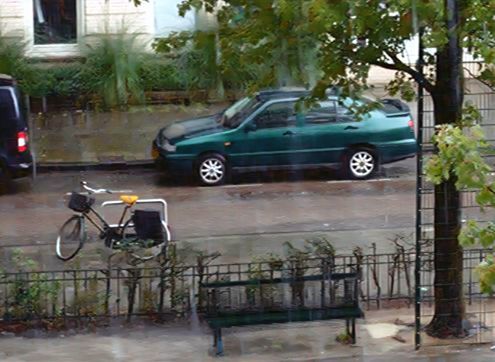}
		\end{subfigure}
		\begin{subfigure}{0.33\linewidth}
			\includegraphics[width=1\linewidth,height=4cm]{3-ours.jpg}
		\end{subfigure}
	\end{center}
	\caption{Visual comparison. The upper row contains the input, results by GMM and DDN, while the lower row contains the results by JORDER, ID-CGAN and our DDC-Net, respectively. Please zoom-in the results to see more details and differences.}
	\label{fig:real1}
	
	\begin{center}
		\begin{subfigure}{0.33\linewidth}
			\includegraphics[width=1\linewidth,height=4cm]{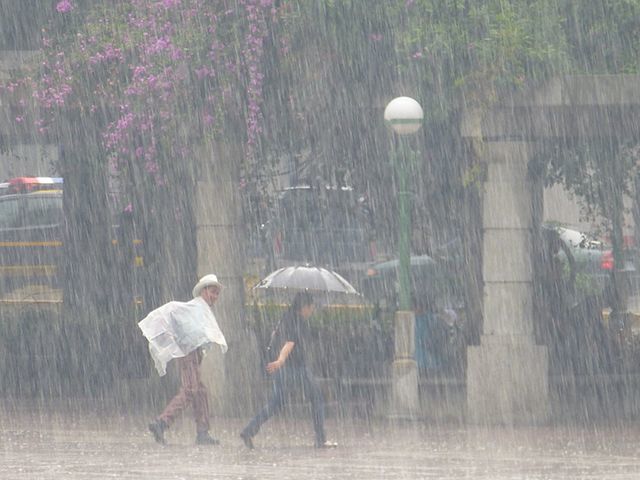}
		\end{subfigure}
		\begin{subfigure}{0.33\linewidth}
			\includegraphics[width=1\linewidth,height=4cm]{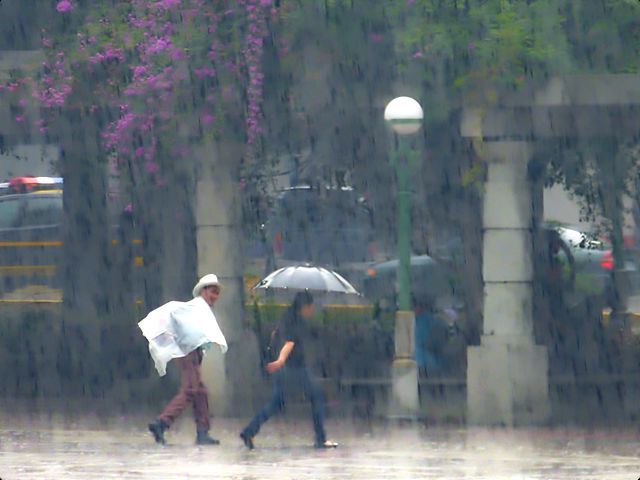}
		\end{subfigure}
		\begin{subfigure}{0.33\linewidth}
			\includegraphics[width=1\linewidth,height=4cm]{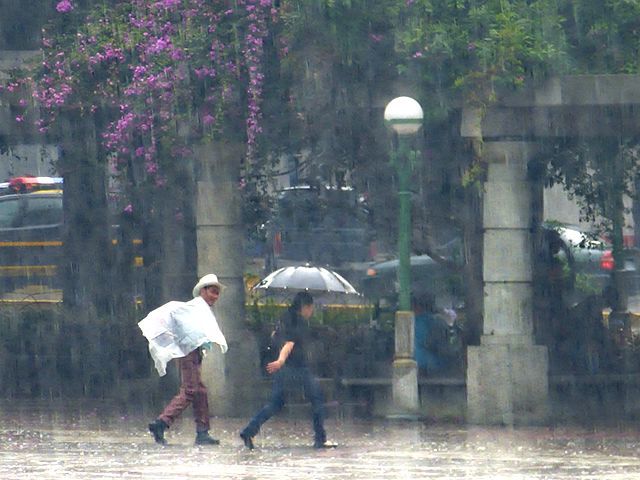}
		\end{subfigure}\\ \vspace{2pt}
		
		\begin{subfigure}{0.33\linewidth}
			\includegraphics[width=1\linewidth,height=4cm]{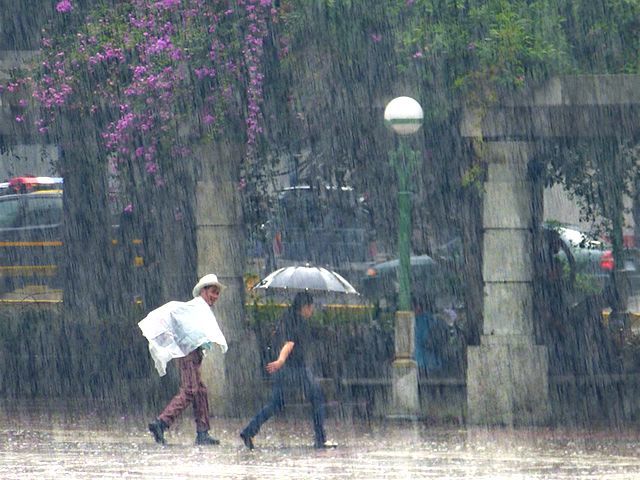}
		\end{subfigure}
		\begin{subfigure}{0.33\linewidth}
			\includegraphics[width=1\linewidth,height=4cm]{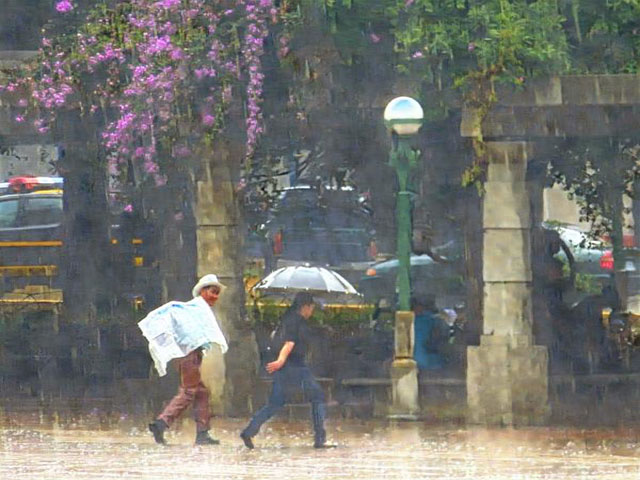}
		\end{subfigure}
		\begin{subfigure}{0.33\linewidth}
			\includegraphics[width=1\linewidth,height=4cm]{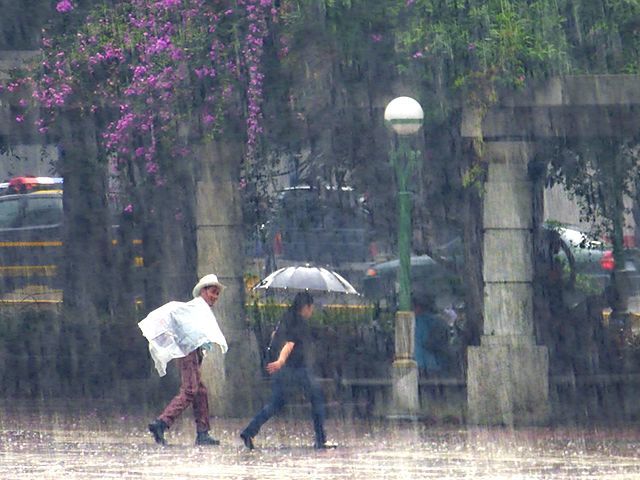}
		\end{subfigure}
	\end{center}
	\caption{Visual comparison. The upper row contains the input, results by GMM and DDN, while the lower row contains the results by JORDER, ID-CGAN and our DDC-Net, respectively. Please zoom-in the results to see more details and differences.}
	\label{fig:real2}
\end{figure*}

\begin{figure*}[t]
	\begin{center}
		\begin{subfigure}{0.33\linewidth}
			\includegraphics[width=1\linewidth,height=4cm]{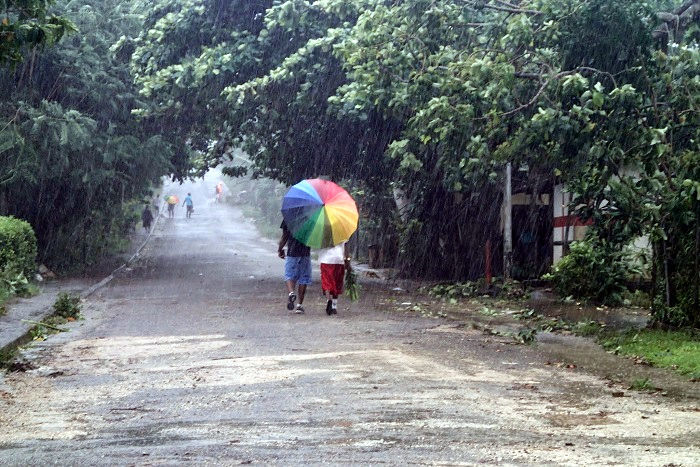}
		\end{subfigure}
		\begin{subfigure}{0.33\linewidth}
			\includegraphics[width=1\linewidth,height=4cm]{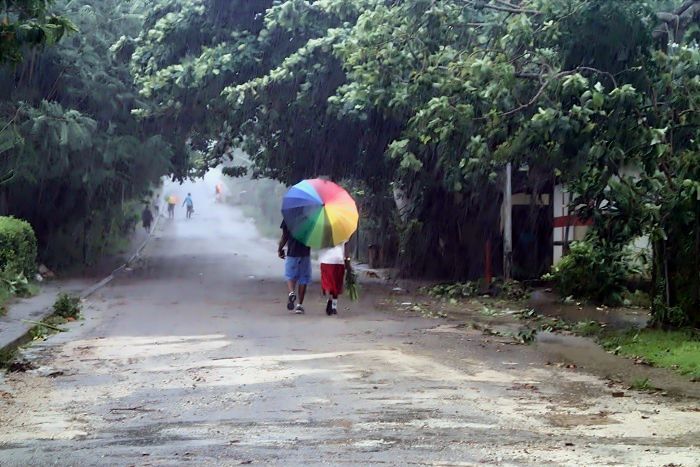}
		\end{subfigure}
		\begin{subfigure}{0.33\linewidth}
			\includegraphics[width=1\linewidth,height=4cm]{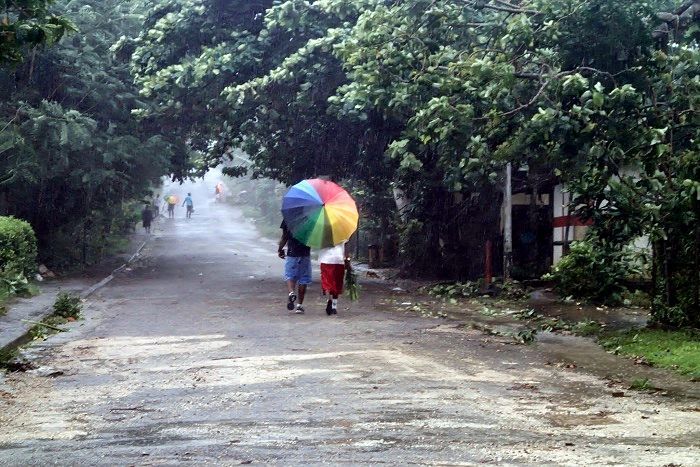}
		\end{subfigure}\\ \vspace{2pt}
		
		\begin{subfigure}{0.33\linewidth}
			\includegraphics[width=1\linewidth,height=4cm]{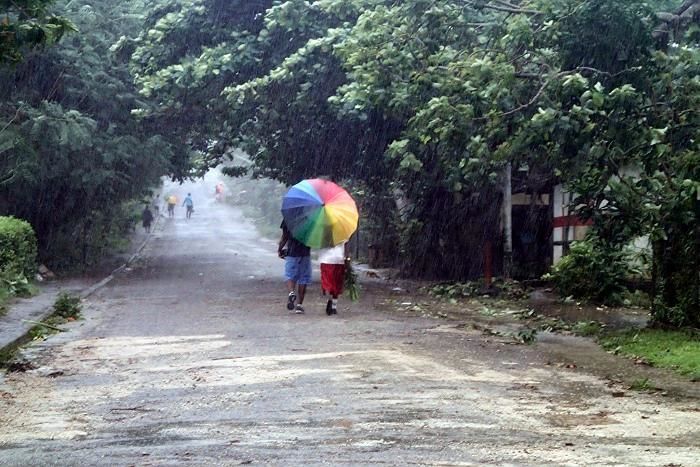}
		\end{subfigure}
		\begin{subfigure}{0.33\linewidth}
			\includegraphics[width=1\linewidth,height=4cm]{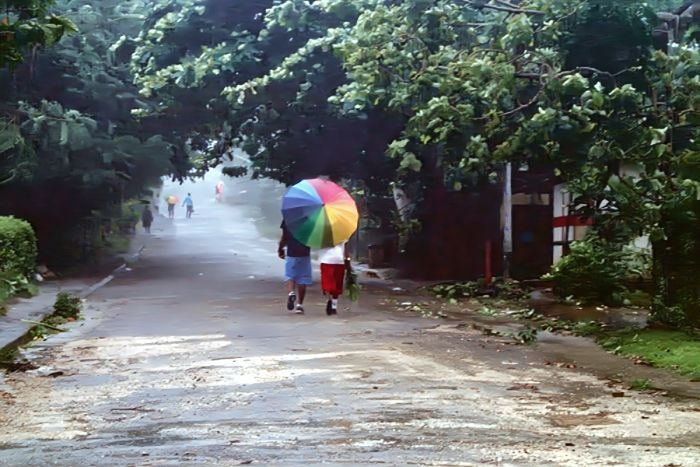}
		\end{subfigure}
		\begin{subfigure}{0.33\linewidth}
			\includegraphics[width=1\linewidth,height=4cm]{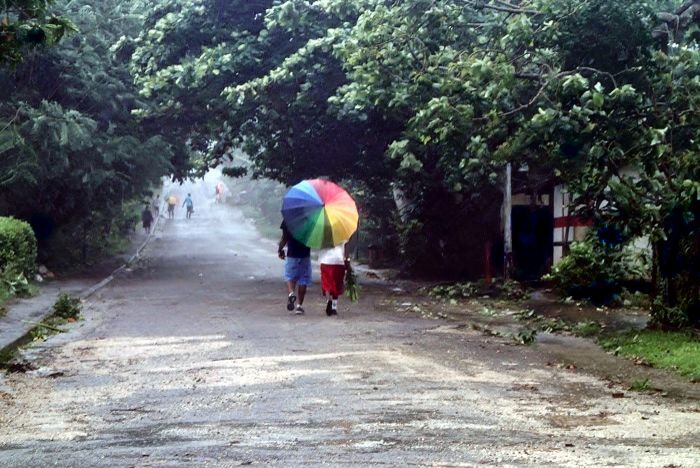}
		\end{subfigure}
	\end{center}
	\caption{Visual comparison. The upper row contains the input, results by GMM and DDN, while the lower row contains the results by JORDER, ID-CGAN and our DDC-Net, respectively. Please zoom-in the results to see more details and differences.}
	\label{fig:real3}
	
	\begin{center}
		\begin{subfigure}{0.33\linewidth}
			\includegraphics[width=1\linewidth,height=4cm]{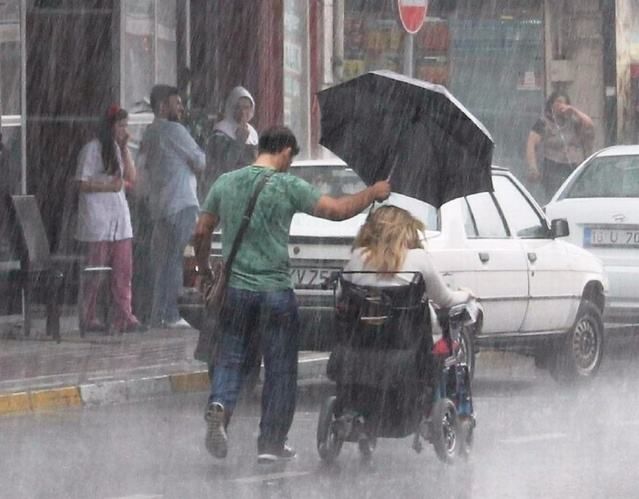}
		\end{subfigure}
		\begin{subfigure}{0.33\linewidth}
			\includegraphics[width=1\linewidth,height=4cm]{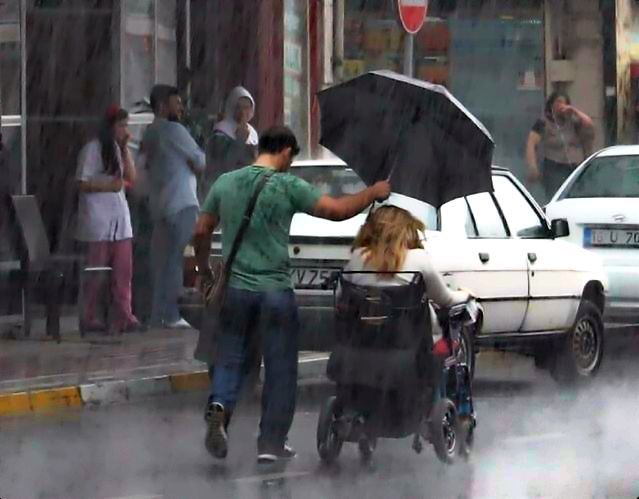}
		\end{subfigure}
		\begin{subfigure}{0.33\linewidth}
			\includegraphics[width=1\linewidth,height=4cm]{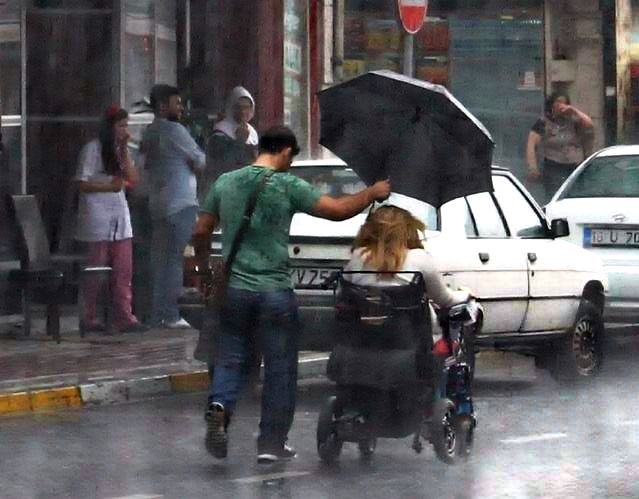}
		\end{subfigure}\\ \vspace{2pt}
		
		\begin{subfigure}{0.33\linewidth}
			\includegraphics[width=1\linewidth,height=4cm]{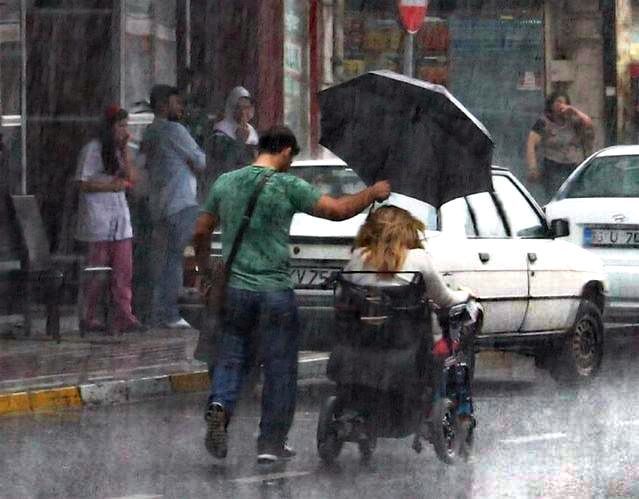}
		\end{subfigure}
		\begin{subfigure}{0.33\linewidth}
			\includegraphics[width=1\linewidth,height=4cm]{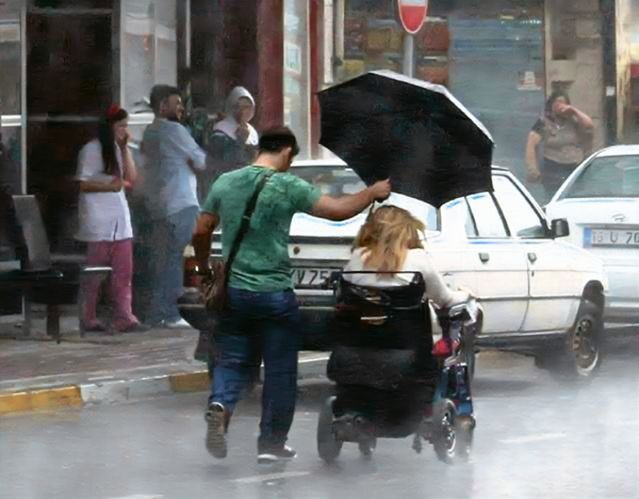}
		\end{subfigure}
		\begin{subfigure}{0.33\linewidth}
			\includegraphics[width=1\linewidth,height=4cm]{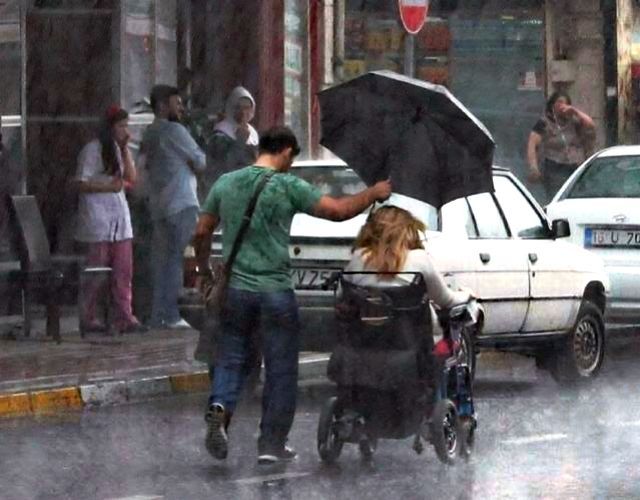}
		\end{subfigure}
	\end{center}
	\caption{Visual comparison. The upper row contains the input, results by GMM and DDN, while the lower row contains the results by JORDER, ID-CGAN and our DDC-Net, respectively. Please zoom-in the results to see more details and differences.}
	\label{fig:real4}
\end{figure*}

\subsection{Synthetic Data}

We first synthesize $7$ rainy images, which are shown in the top row of Fig. \ref{fig:synt}. Table \ref{tab:synm} lists the values of PSNR and SSIM of all the competitors. From the numerical results, we can see that, except for the first case (PSNR and SSIM) and the fifth case (SSIM only), our DDC wins over the others by large margins. While, in the first case, DDC slightly falls behind DDN by about $0.33$ in PSNR and $0.07$ in SSIM, but is still superior to the rest techniques. Figure \ref{fig:synt} provides two visual comparisons between the methods, from which we can observe that our DDC can produce very striking results. GMM leaves rain streaks in both the outputs. For ID-CGAN, it not only suffers from ineffectiveness in rain removal but also alters the color, making the results unrealistic. DDN performs reasonably well for the upper image but not for the lower one, while JORDER produces a good result for the lower image but unsatisfactory for the upper one. In addition, with the composition net disabled (denoted as Ours w/o CN), the decomposition net can still largely separate the rain effect from the inputs. By the complete DDC-Net, the results are further boosted and quite close to the ground truth. Please zoom-in to see more details. This verifies the rationality and effectiveness of our design. One may wonder if only employing the rain branch with the clean background one disabled can produce reasonable results. Our answer is negative, because the rain information is of (relatively) simple pattern and occupies a small fraction of images, merely recovering the rain is not able to guarantee the quality of background recovery. In this paper, we do not explicitly provide such an ablation study.  Besides, the running time is another important aspect to test. Table \ref{tab:time} reports the time taken by different methods. The numbers are averaged over 10 images. We note that as the GMM is implemented using CPUs while the current version of ID-CGAN is only available for GPUs, so we only provide the CPU time for GMM and the GPU time for ID-CGAN. From the table, the evidence demonstrates that our DDC-Net is significantly faster than the competitors, making it more attractive for practice use. It is worth to mention that, GMM requires to execute complex optimization for each input image, while the deep methods including JORDER, DDN, ID-CGAN and our DDC-Net only needs simple feed-forward operations. This is the reason that GMM takes the last place in this competition.

\subsection{Real Data}
The performance of the proposed method is also evaluated on several real-world images. We show the de-raining results in Figs. \ref{fig:real1},\ref{fig:real2},\ref{fig:real3} and \ref{fig:real4}. From the results by GMM \cite{LiS}, we see that the rain streaks are not well removed in the second case, while the rest three cases seem to be over-derained. Regarding the results by JORDER \cite{JORDER}, the under-deraining problem always happens in all the four cases, the rain effect, although alleviated, still obviously exists. The main drawback of ID-CGAN \cite{ID-CGAN} exposed in the comparison is the severe color alteration/degradation after deraining. Furthermore, some halo artifacts emerge, for instance, the region around the right arm of the man in Fig. \ref{fig:real4}.  Arguably, the top $2$ places in this competition should go to DDN \cite{DDN} and our DDC-Net. By taking a closer look at the results, we find that DDN leaves some rain effect on the tree region in the first case and smooths out the texture details, for instance, the earphone wire of the rightmost women, in the fourth case. We notice that for the cases in Fig. \ref{fig:real2} and \ref{fig:real4}, a fast dehazing technique, say LIME \cite{LIME}, is applied to the derained results by GMM, JORDER, DDN and our DDC-Net, while ID-CGAN itself can do the job simultaneously but with a high risk of color degradation. We again emphasize that DDC-Net is about 5 times faster than DDN.

\begin{figure}[t]
	\begin{center}
		\begin{subfigure}{0.48\linewidth}
			\includegraphics[width=1\linewidth]{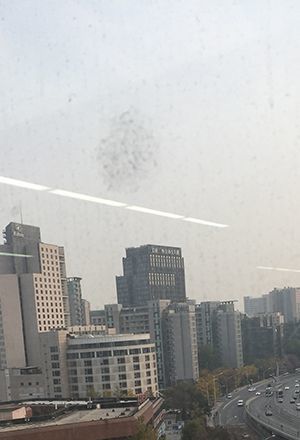}
		\end{subfigure}
		\begin{subfigure}{0.48\linewidth}
			\includegraphics[width=1\linewidth]{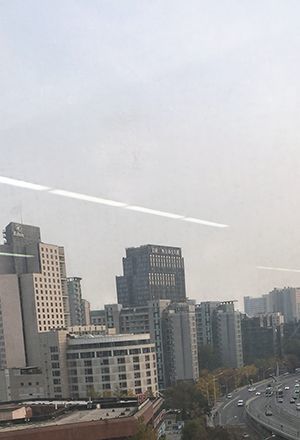}
		\end{subfigure}\\ \vspace{3pt}
		
		\begin{subfigure}{0.48\linewidth}
			\includegraphics[width=1\linewidth]{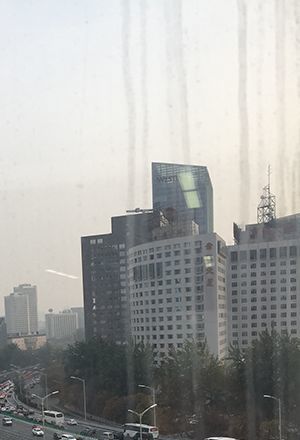}
		\end{subfigure}
		\begin{subfigure}{0.48\linewidth}
			\includegraphics[width=1\linewidth]{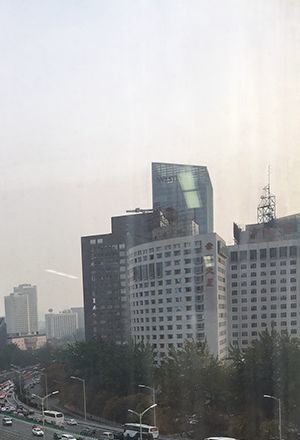}
		\end{subfigure}
	\end{center}
	\caption{Left column: Two natural images interfered by dust/water-based stains on windows. Right column: Corresponding recovered results by our method. }
	\label{fig:dust}
\end{figure}

\section{Discussion and Conclusion}
 Although in this paper we concentrate on the rain removal task, our architecture is general to be applied to other layer-decomposition problems like dust removal and recovery tasks like inpainting. For verifying this point, we show two dust removal results by our DDC-Net as given in Fig. \ref{fig:dust}. From the results, we can observe that the dust/water-based stains on windows are removed effectively and the recovered results are visually striking. As for the efficiency, our design can be further boosted by deep neural network compression techniques, which would highly likely fulfill the real-time requirement in practice. For the composition network, besides the convolution operator, there might be more suitable manners in terms of blending. For example, the so-called Porter Duff operators define $12$ modes including both linear and non-linear blending, and some might be learnable with appropriate parameterization.

Finally, we come to conclude our work. This paper has proposed a novel deep architecture for single image rain removal, namely the deep decomposition-composition network, which consists of two main sub-nets including the decomposition network and the composition network. The decomposition network is built to split rain images into clean background and rain layers. Different from previous architectures, our decomposition model consists of, besides a component representing the desired clean image, an extra component for the rain layer. During the training phase, the additional composition structure is employed to reproduce the input by the separated clean image and rain information for further boosting the quality of decomposition. 
Moreover, our pre-trained model by synthetic data is further fine-tuned by unpaired supervision to have a better adaptation for real cases.
Experimental results on both synthetic and real images have been conducted to reveal the efficacy of our design, and demonstrate its clear advantages in comparison with other state-of-the-art methods.  In terms of running time, our method is significantly faster that the other techniques, which can broaden the applicability of deraining to the tasks with the high-speed requirement. 

\section*{Acknowledgment}
We would like to thank the authors of GMM \cite{LiS}, JORDER \cite{JORDER}, DDN \cite{DDN} and ID-CGAN \cite{ID-CGAN} for sharing their codes.

%
%

\bibliographystyle{ACM-Reference-Format}
\bibliography{sigproc} 


\begin{thebibliography}{00}


\ifx \showCODEN    \undefined \def \showCODEN     #1{\unskip}     \fi
\ifx \showDOI      \undefined \def \showDOI       #1{{\tt DOI:}\penalty0{#1}\ }
  \fi
\ifx \showISBNx    \undefined \def \showISBNx     #1{\unskip}     \fi
\ifx \showISBNxiii \undefined \def \showISBNxiii  #1{\unskip}     \fi
\ifx \showISSN     \undefined \def \showISSN      #1{\unskip}     \fi
\ifx \showLCCN     \undefined \def \showLCCN      #1{\unskip}     \fi
\ifx \shownote     \undefined \def \shownote      #1{#1}          \fi
\ifx \showarticletitle \undefined \def \showarticletitle #1{#1}   \fi
\ifx \showURL      \undefined \def \showURL       #1{#1}          \fi
\providecommand\bibfield[2]{#2}
\providecommand\bibinfo[2]{#2}
\providecommand\natexlab[1]{#1}
\providecommand\showeprint[2][]{arXiv:#2}

\bibitem[\protect\citeauthoryear{Cai, Xu, Jia, Qing, and Tao}{Cai
  et~al\mbox{.}}{2016}]%
        {DehNet}
\bibfield{author}{\bibinfo{person}{B. Cai}, \bibinfo{person}{X. Xu},
  \bibinfo{person}{K. Jia}, \bibinfo{person}{C. Qing}, {and}
  \bibinfo{person}{D. Tao}.} \bibinfo{year}{2016}\natexlab{}.
\newblock \showarticletitle{An end-to-end system for single image haze
  removal}.
\newblock \bibinfo{journal}{{\em IEEE TIP\/}} \bibinfo{volume}{25},
  \bibinfo{number}{11} (\bibinfo{year}{2016}), \bibinfo{pages}{5187--5198}.
\newblock


\bibitem[\protect\citeauthoryear{Chen and Hsu}{Chen and Hsu}{2013}]%
        {ChenS}
\bibfield{author}{\bibinfo{person}{Y. Chen} {and} \bibinfo{person}{C. Hsu}.}
  \bibinfo{year}{2013}\natexlab{}.
\newblock \showarticletitle{A generalized low-rank appearance model for
  spatio-temporally correlated rain streaks}. In \bibinfo{booktitle}{{\em
  ICCV}}. \bibinfo{pages}{1968--1975}.
\newblock


\bibitem[\protect\citeauthoryear{Dong, Chen, He, and Tang}{Dong
  et~al\mbox{.}}{2016}]%
        {PAMISR}
\bibfield{author}{\bibinfo{person}{C. Dong}, \bibinfo{person}{C.~L. Chen},
  \bibinfo{person}{K. He}, {and} \bibinfo{person}{X. Tang}.}
  \bibinfo{year}{2016}\natexlab{}.
\newblock \showarticletitle{Image super-resolution using deep convolutional
  networks}.
\newblock \bibinfo{journal}{{\em IEEE TPAMI\/}} \bibinfo{volume}{38},
  \bibinfo{number}{2} (\bibinfo{year}{2016}), \bibinfo{pages}{295--307}.
\newblock


\bibitem[\protect\citeauthoryear{Dong, Deng, Loy, and Tang}{Dong
  et~al\mbox{.}}{2015}]%
        {ICCVCmp}
\bibfield{author}{\bibinfo{person}{C. Dong}, \bibinfo{person}{Y. Deng},
  \bibinfo{person}{C.~C. Loy}, {and} \bibinfo{person}{X. Tang}.}
  \bibinfo{year}{2015}\natexlab{}.
\newblock \showarticletitle{Compression artifacts reduction by a deep
  convolutional network}. In \bibinfo{booktitle}{{\em ICCV}}.
\newblock


\bibitem[\protect\citeauthoryear{Fu, Huang, Zeng, Huang, Ding, and Paisley}{Fu
  et~al\mbox{.}}{2017}]%
        {DDN}
\bibfield{author}{\bibinfo{person}{X. Fu}, \bibinfo{person}{J. Huang},
  \bibinfo{person}{D. Zeng}, \bibinfo{person}{Y. Huang}, \bibinfo{person}{X.
  Ding}, {and} \bibinfo{person}{J. Paisley}.} \bibinfo{year}{2017}\natexlab{}.
\newblock \showarticletitle{Removing rain from single images via a deep detail
  network}. In \bibinfo{booktitle}{{\em CVPR}}. \bibinfo{pages}{1715--1723}.
\newblock


\bibitem[\protect\citeauthoryear{Garg and Nayar}{Garg and Nayar}{2005}]%
        {GargC}
\bibfield{author}{\bibinfo{person}{K. Garg} {and} \bibinfo{person}{S. Nayar}.}
  \bibinfo{year}{2005}\natexlab{}.
\newblock \showarticletitle{What does a camera see rain?}. In
  \bibinfo{booktitle}{{\em ICCV}}. \bibinfo{pages}{1067--1074}.
\newblock


\bibitem[\protect\citeauthoryear{Garg and Nayar}{Garg and Nayar}{2007}]%
        {GargM2}
\bibfield{author}{\bibinfo{person}{K. Garg} {and} \bibinfo{person}{S. Nayar}.}
  \bibinfo{year}{2007}\natexlab{}.
\newblock \showarticletitle{Vision and Rain}.
\newblock \bibinfo{journal}{{\em IJCV\/}} \bibinfo{volume}{75},
  \bibinfo{number}{1} (\bibinfo{year}{2007}), \bibinfo{pages}{3--27}.
\newblock


\bibitem[\protect\citeauthoryear{Goodfellow, Pouget-Abadie, Mirza, Xu,
  Warde-Farley, Ozair, Courville, and Bengio}{Goodfellow et~al\mbox{.}}{2014}]%
        {goodfellow2014generative}
\bibfield{author}{\bibinfo{person}{I. Goodfellow}, \bibinfo{person}{J.
  Pouget-Abadie}, \bibinfo{person}{M. Mirza}, \bibinfo{person}{B. Xu},
  \bibinfo{person}{D. Warde-Farley}, \bibinfo{person}{S. Ozair},
  \bibinfo{person}{A. Courville}, {and} \bibinfo{person}{Y. Bengio}.}
  \bibinfo{year}{2014}\natexlab{}.
\newblock \showarticletitle{Generative adversarial nets}. In
  \bibinfo{booktitle}{{\em NIPS}}. \bibinfo{pages}{2672--2680}.
\newblock


\bibitem[\protect\citeauthoryear{Guo, Li, and Ling}{Guo et~al\mbox{.}}{2017}]%
        {LIME}
\bibfield{author}{\bibinfo{person}{X. Guo}, \bibinfo{person}{Y. Li}, {and}
  \bibinfo{person}{H. Ling}.} \bibinfo{year}{2017}\natexlab{}.
\newblock \showarticletitle{LIME: Low-light Image Enhancement via Illumination
  Map Estimation}.
\newblock \bibinfo{journal}{{\em IEEE TIP\/}} \bibinfo{volume}{26},
  \bibinfo{number}{2} (\bibinfo{year}{2017}), \bibinfo{pages}{982--993}.
\newblock


\bibitem[\protect\citeauthoryear{Huang, Kang, Wang, and Lin}{Huang
  et~al\mbox{.}}{2014}]%
        {HuangS}
\bibfield{author}{\bibinfo{person}{D. Huang}, \bibinfo{person}{L. Kang},
  \bibinfo{person}{Y. Wang}, {and} \bibinfo{person}{C. Lin}.}
  \bibinfo{year}{2014}\natexlab{}.
\newblock \showarticletitle{Self-learning based image decomposition with
  applications to single image denoising}.
\newblock \bibinfo{journal}{{\em IEEE TMM\/}} \bibinfo{volume}{16},
  \bibinfo{number}{1} (\bibinfo{year}{2014}), \bibinfo{pages}{83--93}.
\newblock


\bibitem[\protect\citeauthoryear{Kang, Lin, and Fu}{Kang et~al\mbox{.}}{2012}]%
        {KangS}
\bibfield{author}{\bibinfo{person}{L. Kang}, \bibinfo{person}{C. Lin}, {and}
  \bibinfo{person}{Y. Fu}.} \bibinfo{year}{2012}\natexlab{}.
\newblock \showarticletitle{Automatic single-image-based rain streaks removal
  via image decomposition}.
\newblock \bibinfo{journal}{{\em IEEE TIP\/}} \bibinfo{volume}{21},
  \bibinfo{number}{4} (\bibinfo{year}{2012}), \bibinfo{pages}{1742--1755}.
\newblock


\bibitem[\protect\citeauthoryear{Kim, Lee, Sim, and Kim}{Kim
  et~al\mbox{.}}{2013}]%
        {KimS}
\bibfield{author}{\bibinfo{person}{J. Kim}, \bibinfo{person}{C. Lee},
  \bibinfo{person}{J. Sim}, {and} \bibinfo{person}{C. Kim}.}
  \bibinfo{year}{2013}\natexlab{}.
\newblock \showarticletitle{Single-image deraining using an adaptive nonlocal
  means filter}. In \bibinfo{booktitle}{{\em ICIP}}. \bibinfo{pages}{914--917}.
\newblock


\bibitem[\protect\citeauthoryear{Li, Tan, Guo, Lu, and Brown}{Li
  et~al\mbox{.}}{2016}]%
        {LiS}
\bibfield{author}{\bibinfo{person}{Y. Li}, \bibinfo{person}{R. Tan},
  \bibinfo{person}{X. Guo}, \bibinfo{person}{J. Lu}, {and} \bibinfo{person}{M.
  Brown}.} \bibinfo{year}{2016}\natexlab{}.
\newblock \showarticletitle{Rain Streak Removal Using Layer Priors}. In
  \bibinfo{booktitle}{{\em CVPR}}.
\newblock


\bibitem[\protect\citeauthoryear{Luo, Xu, and Ji}{Luo et~al\mbox{.}}{2015}]%
        {LuoS}
\bibfield{author}{\bibinfo{person}{Y. Luo}, \bibinfo{person}{Y. Xu}, {and}
  \bibinfo{person}{H. Ji}.} \bibinfo{year}{2015}\natexlab{}.
\newblock \showarticletitle{Removing rain from a single image via
  discriminative sparse coding}. In \bibinfo{booktitle}{{\em ICCV}}.
  \bibinfo{pages}{3397--3405}.
\newblock


\bibitem[\protect\citeauthoryear{Mao, Shen, and Yang}{Mao
  et~al\mbox{.}}{2016}]%
        {mao2016image}
\bibfield{author}{\bibinfo{person}{X. Mao}, \bibinfo{person}{C. Shen}, {and}
  \bibinfo{person}{Y.-B. Yang}.} \bibinfo{year}{2016}\natexlab{}.
\newblock \showarticletitle{Image restoration using very deep convolutional
  encoder-decoder networks with symmetric skip connections}. In
  \bibinfo{booktitle}{{\em NIPS}}.
\newblock


\bibitem[\protect\citeauthoryear{Narasimhan and Nayar}{Narasimhan and
  Nayar}{2003}]%
        {Effect}
\bibfield{author}{\bibinfo{person}{S. Narasimhan} {and} \bibinfo{person}{S.
  Nayar}.} \bibinfo{year}{2003}\natexlab{}.
\newblock \showarticletitle{Contrast restoration of weather degraded images}.
\newblock \bibinfo{journal}{{\em IEEE TPAMI\/}} \bibinfo{volume}{25},
  \bibinfo{number}{6} (\bibinfo{year}{2003}), \bibinfo{pages}{713--724}.
\newblock


\bibitem[\protect\citeauthoryear{Oneata, Revaud, Verbeek, and Schmid}{Oneata
  et~al\mbox{.}}{2014}]%
        {objDet}
\bibfield{author}{\bibinfo{person}{D. Oneata}, \bibinfo{person}{J. Revaud},
  \bibinfo{person}{J. Verbeek}, {and} \bibinfo{person}{C. Schmid}.}
  \bibinfo{year}{2014}\natexlab{}.
\newblock \showarticletitle{Spatio-Temporal Object Detection Proposals}. In
  \bibinfo{booktitle}{{\em ECCV}}. \bibinfo{pages}{737--752}.
\newblock


\bibitem[\protect\citeauthoryear{Pathak, Krahenbuhl, Donahue, Darrell, and
  Efros}{Pathak et~al\mbox{.}}{2016}]%
        {pathak2016context}
\bibfield{author}{\bibinfo{person}{D. Pathak}, \bibinfo{person}{P. Krahenbuhl},
  \bibinfo{person}{J. Donahue}, \bibinfo{person}{T. Darrell}, {and}
  \bibinfo{person}{A. Efros}.} \bibinfo{year}{2016}\natexlab{}.
\newblock \showarticletitle{Context encoders: Feature learning by inpainting}.
  In \bibinfo{booktitle}{{\em CVPR}}.
\newblock


\bibitem[\protect\citeauthoryear{Sun, Pan, and Wang}{Sun et~al\mbox{.}}{2014}]%
        {SunS}
\bibfield{author}{\bibinfo{person}{S. Sun}, \bibinfo{person}{S. Pan}, {and}
  \bibinfo{person}{Y. Wang}.} \bibinfo{year}{2014}\natexlab{}.
\newblock \showarticletitle{Exploiting image structural similarity for single
  image rain removal}. In \bibinfo{booktitle}{{\em ICME}}.
  \bibinfo{pages}{4482--4486}.
\newblock


\bibitem[\protect\citeauthoryear{Tripathi and Mukhopadhyay}{Tripathi and
  Mukhopadhyay}{2014}]%
        {ReviewM}
\bibfield{author}{\bibinfo{person}{A. Tripathi} {and} \bibinfo{person}{S.
  Mukhopadhyay}.} \bibinfo{year}{2014}\natexlab{}.
\newblock \showarticletitle{Removal of rain from videos: a review}.
\newblock \bibinfo{journal}{{\em Signal, Image and Video Processing\/}}
  \bibinfo{volume}{8}, \bibinfo{number}{8} (\bibinfo{year}{2014}),
  \bibinfo{pages}{1421--1430}.
\newblock


\bibitem[\protect\citeauthoryear{Xie, Xu, Chen, Xie, and Xu}{Xie
  et~al\mbox{.}}{2012}]%
        {NIPSDen}
\bibfield{author}{\bibinfo{person}{J. Xie}, \bibinfo{person}{L. Xu},
  \bibinfo{person}{E. Chen}, \bibinfo{person}{J. Xie}, {and}
  \bibinfo{person}{L. Xu}.} \bibinfo{year}{2012}\natexlab{}.
\newblock \showarticletitle{Image denoising and inpainting with deep neural
  networks}. In \bibinfo{booktitle}{{\em NIPS}}. \bibinfo{pages}{341--349}.
\newblock


\bibitem[\protect\citeauthoryear{Xu, Price, Cohen, and Huang}{Xu
  et~al\mbox{.}}{2017}]%
        {xu2017deep}
\bibfield{author}{\bibinfo{person}{N. Xu}, \bibinfo{person}{B. Price},
  \bibinfo{person}{S. Cohen}, {and} \bibinfo{person}{T. Huang}.}
  \bibinfo{year}{2017}\natexlab{}.
\newblock \showarticletitle{Deep Image Matting}.
\newblock \bibinfo{journal}{{\em arXiv preprint arXiv:1703.03872\/}}
  (\bibinfo{year}{2017}).
\newblock


\bibitem[\protect\citeauthoryear{Yang, Tan, Feng, Liu, Guo, and Yan}{Yang
  et~al\mbox{.}}{2017a}]%
        {JORDER}
\bibfield{author}{\bibinfo{person}{W. Yang}, \bibinfo{person}{R.~T. Tan},
  \bibinfo{person}{J. Feng}, \bibinfo{person}{J. Liu}, \bibinfo{person}{Z.
  Guo}, {and} \bibinfo{person}{S. Yan}.} \bibinfo{year}{2017}\natexlab{a}.
\newblock \showarticletitle{Deep joint rain detection and removal from a single
  image}. In \bibinfo{booktitle}{{\em CVPR}}. \bibinfo{pages}{1357--1366}.
\newblock


\bibitem[\protect\citeauthoryear{Yang, Tan, Feng, Liu, Guo, and Yan}{Yang
  et~al\mbox{.}}{2017b}]%
        {yang2017deep}
\bibfield{author}{\bibinfo{person}{W. Yang}, \bibinfo{person}{R.~T. Tan},
  \bibinfo{person}{J. Feng}, \bibinfo{person}{J. Liu}, \bibinfo{person}{Z.
  Guo}, {and} \bibinfo{person}{S. Yan}.} \bibinfo{year}{2017}\natexlab{b}.
\newblock \showarticletitle{Deep Joint Rain Detection and Removal from a Single
  Image}. \bibinfo{pages}{1357--1366}.
\newblock


\bibitem[\protect\citeauthoryear{Yang, Xu, and Luo}{Yang et~al\mbox{.}}{2018}]%
        {yang2018towards}
\bibfield{author}{\bibinfo{person}{X. Yang}, \bibinfo{person}{Z. Xu}, {and}
  \bibinfo{person}{J. Luo}.} \bibinfo{year}{2018}\natexlab{}.
\newblock \showarticletitle{Towards Perceptual Image Dehazing by Physics-based
  Disentanglement and Adversarial Training}.
\newblock  (\bibinfo{year}{2018}).
\newblock


\bibitem[\protect\citeauthoryear{You, Tan, Kawakami, and Ikeuchi}{You
  et~al\mbox{.}}{2013}]%
        {YouM}
\bibfield{author}{\bibinfo{person}{S. You}, \bibinfo{person}{R. Tan},
  \bibinfo{person}{R. Kawakami}, {and} \bibinfo{person}{K. Ikeuchi}.}
  \bibinfo{year}{2013}\natexlab{}.
\newblock \showarticletitle{Adherent raindrop detection and removal in video}.
  In \bibinfo{booktitle}{{\em CVPR}}. \bibinfo{pages}{1035--1042}.
\newblock


\bibitem[\protect\citeauthoryear{Zhang, Sindagi, and Patel}{Zhang
  et~al\mbox{.}}{2017}]%
        {ID-CGAN}
\bibfield{author}{\bibinfo{person}{H. Zhang}, \bibinfo{person}{V. Sindagi},
  {and} \bibinfo{person}{V.~M. Patel}.} \bibinfo{year}{2017}\natexlab{}.
\newblock \showarticletitle{Image De-raining Using a Conditional Generative
  Adversarial Network}.
\newblock \bibinfo{journal}{{\em arXiv:1701.05957v2\/}} (\bibinfo{year}{2017}).
\newblock


\bibitem[\protect\citeauthoryear{Zhang, Zhang, and Yang}{Zhang
  et~al\mbox{.}}{2014}]%
        {Tracking}
\bibfield{author}{\bibinfo{person}{K. Zhang}, \bibinfo{person}{L. Zhang}, {and}
  \bibinfo{person}{M. Yang}.} \bibinfo{year}{2014}\natexlab{}.
\newblock \showarticletitle{Real-Time Compressive Tracking}. In
  \bibinfo{booktitle}{{\em ECCV}}. \bibinfo{pages}{866--879}.
\newblock


\bibitem[\protect\citeauthoryear{Zhang, Zuo, Chen, Meng, and Zhang}{Zhang
  et~al\mbox{.}}{2017}]%
        {TIPDen}
\bibfield{author}{\bibinfo{person}{K. Zhang}, \bibinfo{person}{W. Zuo},
  \bibinfo{person}{Y. Chen}, \bibinfo{person}{D. Meng}, {and}
  \bibinfo{person}{L. Zhang}.} \bibinfo{year}{2017}\natexlab{}.
\newblock \showarticletitle{Beyond a gaussian denoiser: Residual learning of
  deep CNN for image denoising}.
\newblock \bibinfo{journal}{{\em IEEE TIP\/}} \bibinfo{volume}{26},
  \bibinfo{number}{7} (\bibinfo{year}{2017}), \bibinfo{pages}{3142--3155}.
\newblock


\bibitem[\protect\citeauthoryear{Zhang, Li, Qi, Leow, and Ng}{Zhang
  et~al\mbox{.}}{2006}]%
        {ZhangM}
\bibfield{author}{\bibinfo{person}{X. Zhang}, \bibinfo{person}{H. Li},
  \bibinfo{person}{Y. Qi}, \bibinfo{person}{W. Leow}, {and} \bibinfo{person}{T.
  Ng}.} \bibinfo{year}{2006}\natexlab{}.
\newblock \showarticletitle{Rain removal in video by combining temporal and
  chromatic properties}. In \bibinfo{booktitle}{{\em ICME}}.
  \bibinfo{pages}{461--464}.
\newblock


\end{thebibliography}

\end{document}